\documentclass[10pt,twocolumn,letterpaper]{article}

\usepackage{iccv}
\usepackage{times}
\usepackage{epsfig}
\usepackage{graphicx}
\usepackage{amsmath}
\usepackage{amssymb}
\usepackage{booktabs}

\usepackage[pagebackref=true,breaklinks=true,letterpaper=true,colorlinks,bookmarks=false]{hyperref}

\iccvfinalcopy 

\def\iccvPaperID{3020} 

\ificcvfinal\fi

\begin{document}

\title{Object as Query: Lifting any 2D Object Detector to 3D Detection}

\author{\stepcounter{footnote}\normalsize{Zitian~Wang$^{1}$}
\qquad \normalsize{Zehao~Huang$^{2}$} \qquad \normalsize{Jiahui~Fu$^1$} \qquad \normalsize{Naiyan~Wang$^2$} \qquad \normalsize{Si~Liu$^{1}$} \\
    \small{$^{1}$Institute of Artificial Intelligence, Beihang University} \\
	\small{$^{2}$TuSimple} \\
	{\small\tt $\{$wangzt.kghl,zehaohuang18,winsty$\}$@gmail.com} \ \ {\small\tt $\{$jiahuifu,liusi$\}$@buaa.edu.cn}
}

\maketitle
\ificcvfinal\thispagestyle{empty}\fi

\begin{abstract}
3D object detection from multi-view images has drawn much attention over the past few years.
Existing methods mainly establish 3D representations from multi-view images and adopt a dense detection head for object detection, or employ object queries distributed in 3D space to localize objects.
In this paper, we design Multi-View 2D Objects guided 3D Object Detector (MV2D), which can lift any 2D object detector to multi-view 3D object detection. 
Since 2D detections can provide valuable priors for object existence, MV2D exploits 2D detectors to generate object queries conditioned on the rich image semantics. 
These dynamically generated queries help MV2D to recall objects in the field of view and show a strong capability of localizing 3D objects.
For the generated queries, we design a sparse cross attention module to force them to focus on the features of specific objects, which suppresses interference from noises. 
The evaluation results on the nuScenes dataset demonstrate the dynamic object queries and sparse feature aggregation can promote 3D detection capability. 
MV2D also exhibits a state-of-the-art performance among existing methods. 
We hope MV2D can serve as a new baseline for future research. 
Code is available at \url{https://github.com/tusen-ai/MV2D}.

\end{abstract}

\section{Introduction}
\label{sec:intro}
Camera-based 3D object detection in unconstrained real-world scenes has drawn much attention over the past few years. 
Early monocular 3D object detection methods \cite{mousavian20173d,DBLP:journals/corr/abs-1904-07850,DBLP:conf/iccv/Brazil019,DBLP:conf/iccv/SimonelliBPLK19, DBLP:conf/cvpr/KuPW19,FCOS3D, PGD}
typically build their framework following the 2D object detection pipeline. The 3D location and attributes of objects are directly predicted from a single view image.
Though these methods have achieved great progress, they are incapable of utilizing the geometric configuration of surrounding cameras and multi-view image correspondences, which are pivotal for the 3D position of objects in the real world.
Moreover, adapting these methods to multi-view setting relies on sophisticated cross-camera post-processing, which further causes degraded efficiency and efficacy. To handle these problems, recent researchers \cite{DETR3D, BEVDet, BEVFormer, BEVDepth, PETR} propose to directly localize objects in 3D world space based on multi-view images, providing a new paradigm for vision-based 3D object detection.

\begin{figure}[t!]
	\centering
	\vspace{-2mm}
	\includegraphics[width=0.5\textwidth]{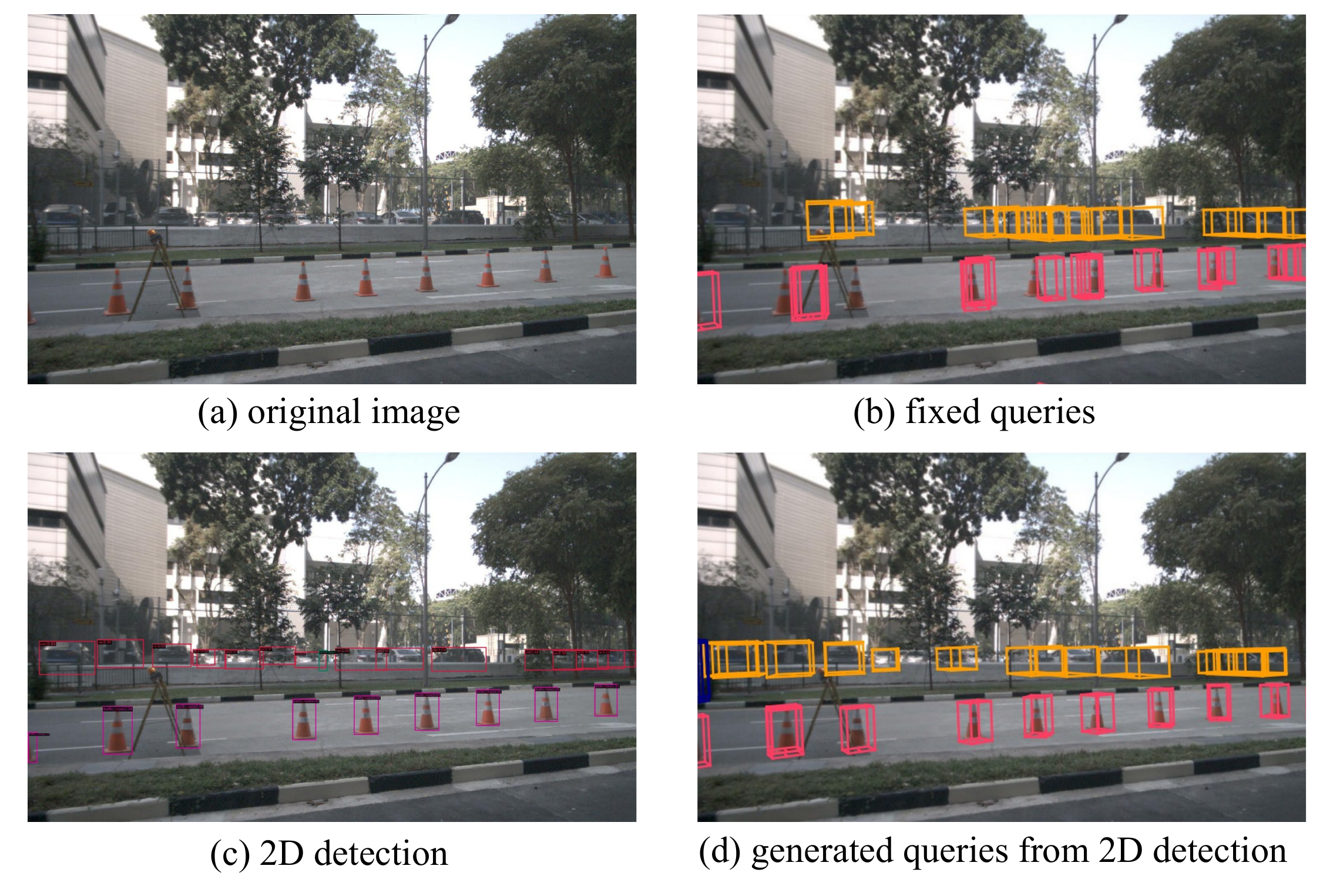}
	\vspace{-6mm}
	\caption{Motivation of MV2D. The 3D detector with fixed object queries (fixed queries means the queries are invariant for different inputs) might mislocate or ignore some objects (b), which are however successfully detected by a 2D detector (c). If generating object queries based on 2D detector, a 3D detector can produce more precise locations (d).}
	\label{fig:figure1}
	\vspace{-5mm}
\end{figure}

According to the representation of fused features, current multi-view 3D object detection methods can be mainly divided into two streams: dense 3D methods and sparse query methods. 
Concretely, dense 3D methods render multi-view features into 3D space, such as Bird’s-Eye-View~(BEV) feature space \cite{BEVDet, BEVDepth, BEVFormer, PolarDETR} or voxel feature space \cite{ImVoxelNet, UVTR}.
However, since the computational costs are squarely proportional to the range of 3D space, they inevitably cannot scale up to large-scale scenarios \cite{FSD}.
Alternatively, query-based methods \cite{DETR3D, PETR} adopt learnable object queries to aggregate features from multi-view images and predict object bounding boxes based on query features.
Although fixed number of object queries avoid computational cost exploding with 3D space, the query number and position relied on empirical prior may cause false positive or undetected objects in dynamic scenarios.

In this paper, we seek a more reliable way to localize objects. The motivation comes from the rapid developments of 2D object detection methods \cite{RCNN, FasterRCNN, lin2017feature, FCOS} which can generate high-quality 2D bounding boxes for object localization in image space.
One natural idea is to turn each 2D detection into one reference for the following 3D detection task. In this way, we design Multi-View 2D Objects guided 3D Object Detector (MV2D), which \emph{could lift any 2D detector to 3D detection, and the 3D detector could harvest all the advancements from 2D detection field.}

Given the input multi-view images, we first obtain 2D detection results from a 2D detector and then generate a dynamic object query for each 2D bounding box.
Instead of aggregating features from all regions in the multi-view inputs, one object query is forced to focus on one specific object.
To this end, we propose an efficient relevant feature selection method based on the 2D detection results and camera configurations.
Then the dynamically generated object queries, together with their 3D position embedded relevant features, are input to a transformer decoder with sparse cross-attention layers.
Lastly, the updated object queries predict the final 3D bounding boxes.

Thanks to the powerful 2D detection performances, the dynamic queries generated from 2D detection results can cover most objects that appeared in the images, leading to higher precision and recall, especially for small and distant objects compared with fixed queries. 
As shown in Figure~\ref{fig:figure1}, 
the objects that a fixed query-based 3D detector might miss can be recalled in MV2D with the help of a 2D object detector.
Theoretically, since 3D object queries stem from 2D detection results, our method can benefit from all off-the-shelf excellent 2D detector improvements.
Our contributions can be summarized as: 
\begin{itemize}
    \item We propose a framework MV2D that can lift any 2D object detector to multi-view 3D object detection. 
    \item We demonstrate that dynamic object queries and aggregation from certain relevant regions in multi-view images based on 2D detection can boost 3D detection performance.
    \item We evaluate MV2D on the standard nuScenes dataset, and it achieves state-of-the-art performance.
\end{itemize}


\begin{figure*}[t!]
	\centering
	\vspace{-4mm}
	\includegraphics[width=1.0\textwidth]{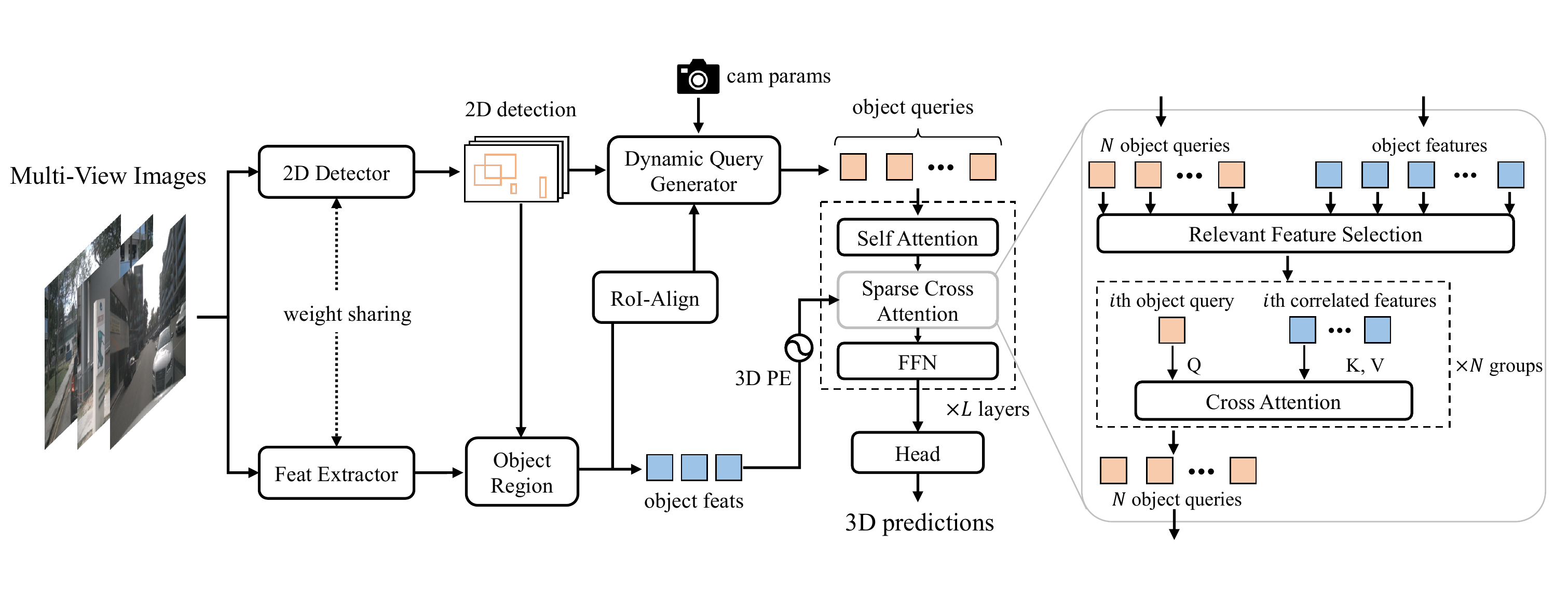}
	\vspace{-12mm}
	\caption{The framework of the proposed MV2D.
	Given the input multi-view images, image feature maps are extracted by a feature extractor. Meanwhile, a 2D detector is used to obtain per-view 2D detection results. 
 Dynamic query generator takes object features, 2D detection boxes and camera parameters as input to initialize a set of object queries. RoI-Align is applied to the object regions to obtain the fixed length object features for query generator.
 All the features fallen in the object regions are decorated with  3D PE (3D position embedding)~\cite{PETR}, then the object queries and object features are input to a decoder to update query features. Compared to  vallina transformer decoder, the decoder in MV2D employs sparse cross attention where each object query only interacts with its relevant features. 
	Lastly, a prediction head is applied to the updated object queries to generate 3D detection results.
}
	\vspace{-4mm}
	\label{fig:framework}
\end{figure*}

\section{Related Work}
\label{sec:related_work}
\subsection{2D Object Detection}
2D object detection aims to predict and localize objects in 2D images, which is a fundamental problem in computer vision.
The RCNN series~\cite{RCNN, FastRCNN, FasterRCNN, lin2017feature} propose a two-stage pipeline for 2D object detection. 
These methods first generate a set of region proposals likely to contain objects in the image, then make a second stage decision on the object classification and bounding box regression.
Another group of researchers investigates to detect objects in a single-stage pipeline~\cite{YOLO, FCOS, RetinaNet}, aiming to provide faster detectors for deployment.
Recently, DETR~\cite{DETR} introduces a transformer encoder-decoder architecture into 2D object detection and formulates object detection as a set prediction problem. In DETR, a set of learnable object queries are adopted to interact with image features and are responsible for detecting objects.
Deformable DETR~\cite{Deformable-DETR} improves DETR by introducing multi-scale deformable attention. Besides, it transfers the vanilla DETR into a two-stage framework, where region proposals are first generated based on encoder features and then sent to the decoder as object queries for further refinement. 
Efficient DETR~\cite{efficientdetr} conducts research on initializing queries for 2D detection, but it does not investigate how to transfer and tailor this initialization to 3D space.

\subsection{Vision-based 3D Object Detection}
The goal of vision-based 3D object detection is to predict 3D bounding boxes from camera images.
Many previous works~\cite{FCOS3D, DBLP:conf/cvpr/KuPW19, DBLP:conf/iccv/Brazil019, DBLP:conf/iccv/ParkAG0G21, DBLP:conf/iccv/SimonelliBPLK19, PGD} perform the task based on single-view image setting.
These methods mostly focus on instance depth estimation and use an extra depth prediction module to complement 2D detectors on depth information.
However, when dealing with multi-view images surrounding the ego vehicle, these methods need to detect objects from each view and project them to the same 3D space with cumbersome NMS post-processing to merge results.

Recently, some works attempt to directly detect objects in 3D space from multi-view images. 
One branch of these methods lift 2D image into 3D space, then conduct detection based on the 3D representations.
ImVoxelNet~\cite{ImVoxelNet} builds a 3D voxelized space and samples image features from multi-view to obtain the voxel representation. BEVFormer~\cite{BEVFormer} leverages dense BEV queries to project and aggregate features from multi-view images by deformable attention~\cite{Deformable-DETR}.
PolarFormer~\cite{jiang2022polarformer} introduces the Polar representation to model BEV space.
BEVDet and BEVDepth ~\cite{BEVDet, BEVDepth} adopt the Lift-Splat module~\cite{LSS} to transform multi-view image features into the BEV representation based on the predicted depth distribution.
Although such 3D space representations are conducive to unifying multi-view images,
the memory consumption and computational cost would increase with the enlarging of the detection range in 3D space~\cite{FSD}.

Another group of works adopts learnable object queries to aggregate image features and predict objects following DETR~\cite{DETR}.
DETR3D~\cite{DETR3D} generates 3D reference points from object queries and projects them to multi-camera images by camera parameters. Then the query features are refined by the corresponding point features sampling from images.
In contrast to establishing fixed mapping through 3D-to-2D query projecting, some methods learn the flexible mapping via attention mechanisms~\cite{CVT, PETR}.
PETR series~\cite{PETR, petrv2} introduce 3D position-aware image features and learns the flexible mapping between query and image features by global cross-attention.
Nonetheless, these approaches usually require dense object queries distributed in 3D space to ensure sufficient recall of objects. 

In this paper, we propose a 2D objects guided framework for multi-view 3D object detection. Based on 2D detectors, our method can count on dynamic queries to recall the objects and eliminate the interference of noises and distractors.


\section{Method}

\subsection{Overview}

The overall pipeline of MV2D is shown in Figure~\ref{fig:framework}.
Given $N$ multi-view images $\mathcal{I} = \{\mathbf{I}_v \mid 0\leq v < N\}$, image feature maps $\mathcal{F} = \{\mathbf{F}_v \mid 0 \leq v< N\}$ are first extracted from the input images using a backbone network, where $\mathbf{F}_v \in \mathbb{R}^{H^{f} \times W^{f} \times C}$ is the extracted feature map from the $v$-th image. 
To get 2D object detection, a 2D object detector, i.e., Faster R-CNN~\cite{FasterRCNN}, is applied to all the input images, resulting in a set of 2D object bounding boxes $\mathcal{B} = \{\mathbf{B}_v \mid 0\leq v < N\}$, where $\mathbf{B}_{v} \in \mathbb{R}^{M_v \times 4}$ represents the predicted 2D bounding boxes in the $v$-th image and $M_v$ is the number of detected boxes. 
In practice, the weights of the backbone of 2D detector can also be used for subsequent feature extraction.
 
Different from common methods in multi-view 3D object detection which adopt fixed object queries across the whole dataset~\cite{DETR3D,PETR}, MV2D generates object queries conditioned on the input multi-view images.
Given the predicted 2D bounding boxes and extracted image features, MV2D employs RoI-Align~\cite{MaskRCNN} to extract fixed length object features.
Then the object features together with corresponding bounding boxes and camera parameters are input to the dynamic object query generator to produce object queries.
For each object query, relevant image features in multi-view images are selected based on 2D detection results and camera parameters through a relevant feature selection module. 
Then the object queries interact with each other and integrate information from relevant object features iteratively through transformer decoder layers~\cite{vaswani2017attention}. Finally, a simple feed-forward network (FFN) head is used to generate 3D object predictions with updated features.

\subsection{Dynamic Object Query Generation}
\label{sec:query_generator}
With the help of an effective 2D object detector, the object existences are well demonstrated and the object locations are constrained within certain image regions, thus providing valuable priors for localizing objects in 3D space. 

\begin{figure}[t!]
	\centering
	\vspace{-5mm}
	\includegraphics[width=0.46\textwidth]{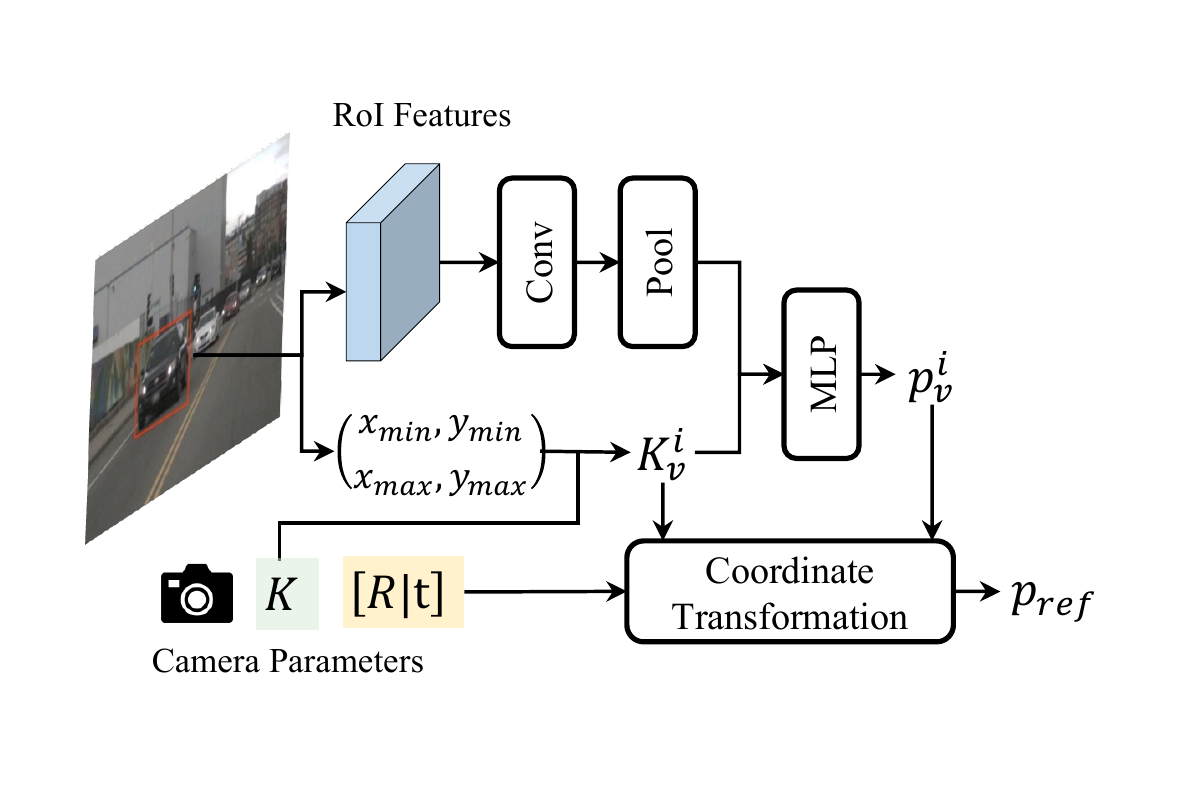}
	\vspace{-6mm}
	\caption{Dynamic object query generator.}
	\label{fig:query_generator}
	\vspace{-5mm}
\end{figure}

To generate object queries from 2D detection, we propose a dynamic query generator as in Figure~\ref{fig:query_generator}. The dynamic query generator derives a 3D reference point $\mathbf{p}_{ref} \in \mathbb{R}^{3}$ in 3D world space for each RoI. Then the object query is generated from $\mathbf{p}_{ref}$ by a positional encoding layer~\cite{vaswani2017attention}.

Specifically, given the 2D object detection results $\mathcal{B}$ and image feature maps $\mathcal{F}$ from all the images, we first extract object RoI features $\mathcal{O}$ through RoI-Align, where $\mathbf{O}_v \in \mathbb{R}^{M_v \times H^{roi} \times W^{roi} \times C}$ are the RoI features corresponding to 2D object bounding boxes $\mathbf{B}_v$:
\begin{equation}
    \mathbf{O}_v = \text{RoI-Align}(\mathbf{F}_v, \mathbf{B}_v), \quad  0 \leq v < N.
\end{equation}
The RoI features $\mathbf{O}_v$ contain sufficient object appearance information to infer the object center locations in image space.
However, further estimating the object depths directly from RoI features is difficult. As the object region in original image space is rescaling into a fixed sized RoI, the geometric information contained in original images is missed.

Taking this into consideration, we apply equivalent camera intrinsic transformation to each RoI, such that the rescaling operation conducted on different RoIs is equivalent to perspective projection with different camera parameters. 
Thus a point in the RoI coordinate system can be transformed into 3D world space with equivalent camera intrinsic:
\begin{equation}
    \mathbf{p}_{ref} = [\mathbf{R}|\mathbf{t}]^{-1} (\mathbf{K}_{v}^i)^{-1}\  \mathbf{p}_{v}^i,
    \label{equ:coordinate_transformation}
\end{equation}
where $\mathbf{p}_{v}^i\in \mathbb{R}^4$ (homogeneous coordinate) represents the 2.5D point in RoI coordinate system, $\mathbf{p}_{ref}\in \mathbb{R}^4$ represents the point in 3D world space, $\mathbf{K}_{v}^i$ is the equivalent camera intrinsic of that RoI and $[\mathbf{R}|\mathbf{t}]$ is the camera extrinsic.

Denote the RoI size is $H^{roi} \times W^{roi}$ (e.g., $7\times7$), the $i$-th 2D object bounding box in the $v$-th view is $\mathbf{B}_v^i=(x_{min}^i,y_{min}^i,x_{max}^i, y_{max}^i)$,
and the original camera intrinsic matrix is:
\begin{equation}
\mathbf{K}_v =
{\small
\begin{bmatrix}
    f_x  & 0 & o_x &  0\\
    0 & f_y & o_y & 0\\
    0 & 0 & 1 & 0 \\
    0 & 0 & 0 & 1
\end{bmatrix}}.
\end{equation}
The equivalent camera intrinsic for $i$-th RoI can be formulated as:
\begin{equation}
    \mathbf{K}_{v}^i = 
    {\small
    \begin{bmatrix}
        f_x * r_x & 0 & (o_x-x_{min}^i)*r_x & 0\\
        0 & f_y * r_y & (o_y-y_{min}^i)*r_y & 0\\
        0 & 0 & 1 & 0 \\
        0 & 0 & 0 & 1
    \end{bmatrix}}
    ,
\end{equation}
where $r_x = W^{roi} / (x_{max}^i - x_{min}^i), r_y = H^{roi} / (y_{max}^i - y_{min}^i)$.

Since the equivalent camera intrinsic matrix contains the geometric property of the camera and object, we adopt a small network to implicitly encode the object location $\mathbf{p}_{v}^i \in \mathbb{R}^4$ based on the RoI feature $\mathbf{o}_v^i$ and the equivalent camera intrinsic $\mathbf{K}_{v}^i$:
\begin{equation}
    \mathbf{p}_{v}^i = \mathcal{H}(\text{MLP}(\text{Pool}(\text{Conv}(\mathbf{o}_v^i));\mathbf{K}_v^i)),
\end{equation}
where $(;)$ means feature concatenation and $\mathcal{H(\cdot)}$ represents homogeneous transformation.

The implicitly encoded object location $\mathbf{p}_{v}^i$, which can be seen as the 3D object location in RoI coordinate system, is then transformed into 3D world space using equivalent camera intrinsic and camera extrinsic as in Equation~\ref{equ:coordinate_transformation}. Consequently, the transformed 3D location serves as reference point $\mathbf{p}_{ref}$ for localizing objects in 3D world space.

By this transformation, MV2D generates a set of dynamic reference points $\mathcal{P}_{ref}=\{\mathbf{P}_{ref, v} \in \mathbb{R}^{M_v \times 3 } \mid 0\leq v < N\}$ based on 2D detections and camera configurations. 
These reference points are then passed through a positional encoding layer~\cite{vaswani2017attention} to
initialize a set of object queries $\mathcal{Q}=\{{\mathbf{Q}_v \in \mathbb{R}^{M_v \times C}} \mid 0\leq v < N\}$. Concretely, each object query $\mathbf{q}$ is obtained by:

\vspace{-6.mm}
\begin{align}
&\mathbf{q}=\text{Linear}(PE), \\
&PE_{[6i:6i+3]}= \sin(\mathbf{p}_{ref}/10000^{2i/C}), 0\leq i < C/2, \\
&PE_{[6i+3:6i+6]}=\cos(\mathbf{p}_{ref}/10000^{2i/C}),0\leq i < C/2.
 \end{align}

\subsection{Relevant Object Feature Selection}
\label{sec:correlated_feature_selection}
Each object is only captured in a sub-region within the multi-view images.
By focusing on the relevant region w.r.t a target object, we can eliminate the distractors and noises that may hinder the object localization performance.  

To this end, we propose to select relevant features for each object query's updating. 
Since 2D object detectors can predict convincing 2D object proposals, 
the detected object bounding boxes imply which region contains the most distinctive information about an object. So we consider two parts of the relevant features of an object: (1) the 2D bounding box $\mathbf{b}_v^i$ from where the object query is generated; (2) the bounding boxes in other views that correspond to $\mathbf{b}_v^i$. 

\begin{figure}[t!]
	\centering
	\includegraphics[width=0.46\textwidth]{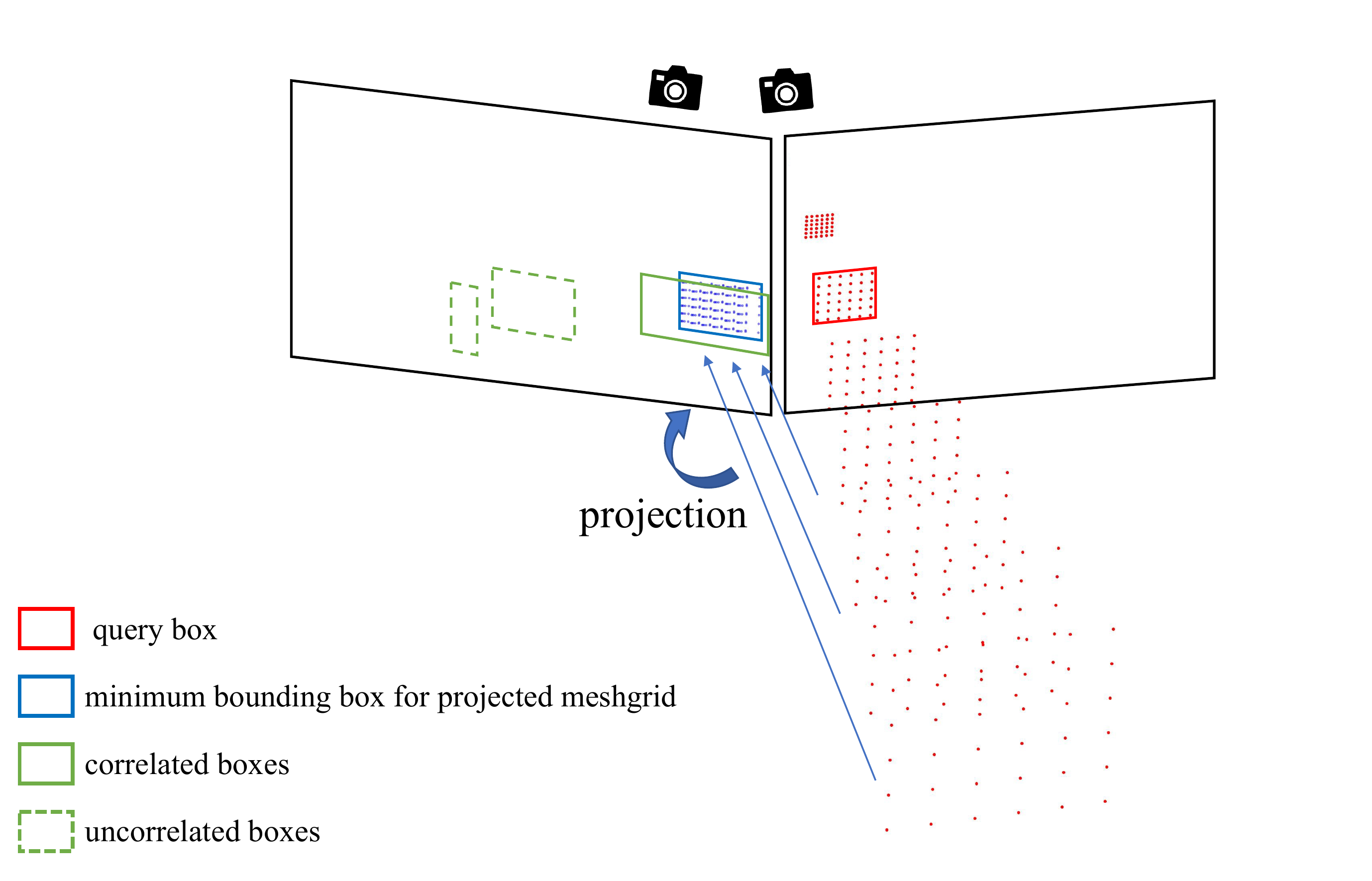}
	\vspace{-0mm}
	\caption{Illustration of relevant region selection. Each query box generates a discretized camera frustum from 3D meshgrid. The camera frustum is then projected to another view's pixel coordinate to calculate a minimum bounding box. Then the relevant box is selected based on the overlap with the minimum bounding box.}
	\label{fig:correlated_feature_selection}
	\vspace{-4mm}
\end{figure}

In this paper, we develop an efficient method to associate the bounding boxes in other views
for a given query.
As shown in Figure~\ref{fig:correlated_feature_selection}, we create a 3D meshgrid $\mathbf{G}\in \mathbb{R}^{W^{roi}\times H^{roi}\times D \times 4}$ for each RoI. We denote $\mathbf{g}_{x,y,z} = (x \times d_z, y \times d_z, d_z, 1)$ as each point in the meshgrid, where $(x, y)$ is the coordinate in the RoI, $d_z \in \{d_0, d_1,\ldots, d_{D-1}\}$ is a set of predifined depth values.
The meshgrid of RoI $\mathbf{b}_v^i$ is transformed into the coordinate system of the $w$-th view:
\begin{equation}
    \mathbf{g}_{v\rightarrow w;x,y,z}^i = \mathbf{K}_w\mathbf{T}_{v\rightarrow w}(\mathbf{K}_v^i)^{-1}\mathbf{g}_{x,y,z},
\end{equation}
where $\mathbf{g}_{v\rightarrow w;x,y,z}^i$ is the transformed meshgrid point in the $w$-th view and $\mathbf{T}_{v\rightarrow w}$ is the coordinate transformation matrix from the $v$-th view to the $w$-th view.
Then the minimum bounding box $\mathbf{b}_{v\rightarrow w}^i$ for the transformed meshgrid $\mathbf{G}_{v\rightarrow w}^i$  can serve as guidance to find the relevant foreground regions for $\mathbf{b}_v^i$.
In this paper, we consider 2 kinds of rules, ``top1 IoU'' and ``all overlapped'', to select relevant foreground regions $\mathcal{R}_v^i$ for $\mathbf{b}_v^i$ from  other views.

\textbf{Top1 IoU:}  the boxes that have the highest Intersection-over-Union (IoU) with $\mathbf{b}_{v\rightarrow w}^i$ (if greater than 0) are selected as relevant regions for $\mathbf{b}_v^i$:
\begin{equation}
    \mathcal{R}_v^i = \left\{
    \mathop{\arg\max}\limits_{\mathbf{b}_w^j \in \mathbf{B}_w} \text{IoU}(\mathbf{b}_{v\rightarrow w}^i, \mathbf{b}_w^j )
    |v\neq w\right\}.
\end{equation}

\textbf{All overlapped:} all the detected boxes that has overlap with  $\mathbf{b}_{v\rightarrow w}^i$ are selected as relevant boxes:
\begin{equation}
    \mathcal{R}_v^i  = \left\{
    \mathbf{b}_w^j| 
    \text{IoU}(\mathbf{b}_{v\rightarrow w}^i, \mathbf{b}_w^j ) > 0, v\neq w
    \right\}.
\end{equation}

Then, all the image features that fall into the region of $\mathbf{b}_v^i$ and $\mathcal{R}_v^i$ are regarded as relevant features and will be adopted to update object query $\mathbf{q}_v^i$. 

\begin{table*}[t]
\begin{center}
\setlength{\tabcolsep}{0.95mm}
\begin{tabular}{l | c c | c c | c c c c c}
        \toprule[1.5pt]
		Method & Backbone & Resolution & NDS$\uparrow$ & mAP$\uparrow$ & mATE$\downarrow$ & mASE$\downarrow$ & mAOE$\downarrow$ & mAVE$\downarrow$ & mAAE$\downarrow$ \\
		\midrule
		DETR3D~\cite{DETR3D} & ResNet-50 & 1600 $\times$ 900 & 0.373 & 0.302 & 0.811 & 0.282 & 0.493 & 0.979 & 0.212\\
        PETR~\cite{PETR} & ResNet-50 & 1408 $\times$ 512 & 0.403 & 0.339 & 0.748 & 0.273 & 0.539 & 0.907 & 0.203\\
        PETRv2~\cite{petrv2} & ResNet-50 & 1600 $\times$ 640 & 0.494 &0.398 &0.690 &0.273 &0.467 &0.424 &0.195\\
        MV2D-S$\ddag$ & ResNet-50 & 1408 $\times$ 512 & 0.440 & 0.398 &  0.665&  0.269&  0.507&  0.946& 0.203\\
        MV2D-T$\ddag$ & ResNet-50 & 1600 $\times$ 640 & \textbf{0.546}&  \textbf{0.459} & \textbf{0.613} & \textbf{0.265} & \textbf{0.388} & \textbf{0.385} & \textbf{0.179}\\ 
        
		\midrule
		FCOS3D$\dag$~\cite{FCOS3D} & ResNet-101 & 1600 $\times$ 900 & 0.415 & 0.343 & 0.725 & 0.263 & 0.422 & 1.292 & \textbf{0.153}\\
        PGD$\dag$~\cite{PGD} & ResNet-101 & 1600 $\times$ 900 & 0.428 & 0.369 & 0.683 & \textbf{0.260} & 0.439 & 1.268 & 0.185\\
		DETR3D$\dag$~\cite{DETR3D} & ResNet-101 & 1600 $\times$ 900 & 0.425 & 0.346 & 0.773 & 0.268 & 0.383 & 0.842 & 0.216\\
        BEVFormer-S$\dag$~\cite{BEVFormer} & ResNet-101 & 1600 $\times$ 900 & 0.448 & 0.375 & 0.725 & 0.272 & 0.391 & 0.802 & 0.200\\
        PETR$\dag$~\cite{PETR} & ResNet-101 & 1600 $\times$ 900 & 0.442 & 0.370 & 0.711 & 0.267 & 0.383 & 0.865 & 0.201\\
        BEVFormer~\cite{BEVFormer}$\dag$& ResNet-101 & 1600 $\times$ 900 & 0.517&0.416 & 0.673 &0.274& 0.372 &0.394 &0.198\\
        PolarFormer~\cite{jiang2022polarformer}$\dag$& ResNet-101 & 1600 $\times$ 900 & 0.528 & 0.432 & 0.648 &0.270 &0.348 &0.409 &0.201\\
        PETRv2$\dag$~\cite{petrv2}& ResNet-101 & 1600 $\times$ 640 &0.524& 0.421& 0.681 &0.267 &0.357 &0.377& 0.186\\
        MV2D-S$\ddag$ & ResNet-101 & 1600 $\times$ 640 & 0.470 & 0.424 &  0.654&  0.267&  0.416&  0.888& 0.200\\
        MV2D-T$\ddag$ & ResNet-101 & 1600 $\times$ 640 & \textbf{0.561} & \textbf{0.471} & \textbf{0.593} &0.262 &\textbf{0.340} &\textbf{0.368} &0.184\\
        
		\bottomrule[1.5pt]
	\end{tabular}
\caption{
\textbf{3D object detection results on nuScenes \texttt{val} set}. $\dag$: the model is initialized from FCOS3D. $\ddag$: the model is pretrained on nuImages. 
}
\label{tab:val_set}
\end{center}
\vspace{-3mm}
\end{table*}

\subsection{Decoder with Sparse Cross Attention}
The generated object queries interact with their relevant features through a DETR-like transformer decoder~\cite{DETR, PETR}.
3D position embedding is provided for the relevant features following PETR~\cite{PETR}. 
The difference in our decoder lies in the sparse cross attention layer as shown in Figure~\ref{fig:framework}.
Instead of using the whole multi-view feature maps to construct keys and values shared by all the queries, MV2D only uses the relevant features to construct their own keys and values for each object query. This design not only contributes a compact set of keys and values
, but also prevents the object queries from being interfered by background noises and distractors.  
Lastly, we apply classification head and regression head composed of MLPs  to the updated object queries to predict the final 3D object detection results.
Note that in 3D object detection, the predicted object center $\hat{\mathbf{b}}_{center}$ is composed of $\mathbf{p}_{ref}$ and offset predicted by query $\hat{\mathbf{b}}_{offset}$, i.e., $\hat{\mathbf{b}}_{center}=\mathbf{p}_{ref}+\hat{\mathbf{b}}_{offset}$.

\subsection{Loss Functions}
In our implementation, the 2D detector and the 3D detector are jointly trained in MV2D, with the backbone weights shared across both detectors.
The 2D object detection loss $\mathcal{L}_{2d}$ is directly taken from 2D detectors.
As for 3D object detection loss, we follow previous works~\cite{DETR3D, PETR} to use Hungarian algorithm~\cite{kuhn1955hungarian} for label assignment. Focal loss~\cite{RetinaNet} and $L_1$ loss are adopted for classification and box regression respectively. The 3D object detection loss can be summarized as:
\begin{equation}
    \mathcal{L}_{3d} = \lambda_{cls3d} \cdot \mathcal{L}_{cls3d} + \mathcal{L}_{reg3d}.
\end{equation}
The overall loss function of MV2D is:
\begin{equation}
    \mathcal{L} = \mathcal{L}_{2d} + \lambda_{3d} \cdot \mathcal{L}_{3d},
\end{equation}
where $\lambda_{3d}$ is the weight term to balance 2D detection and 3D detection losses, set to $0.1$ in our experiments.

\subsection{Multi-Frame Input}
By considering all the input images  (whether from the same timestamp or not) as  camera views with different extrinsic matrices, MV2D can deal with multi-frame input without modification on the pipeline. 
The object queries not only interact with the relevant object features from the same timestamp,  but also the relevant object features from other timestamps. The relevant object feature selection keeps the same as described in Section~\ref{sec:correlated_feature_selection}.


\begin{table*}[t!]
\begin{center}
\setlength{\tabcolsep}{1.20mm}
\begin{tabular}{l |c |c c| c c c c c c }
\toprule[1.5pt]
Method & Backbone & NDS$\uparrow$  & mAP$\uparrow$  & mATE$\downarrow$     & mASE$\downarrow$     & mAOE$\downarrow$ & mAVE$\downarrow$ & mAAE$\downarrow$ \\
\midrule

FCOS3D$\dag$~\cite{FCOS3D}  & ResNet-101 & 0.428 & 0.358 & 0.690 & 0.249 & 0.452 & 1.434 & 0.124 \\
PGD$\dag$~\cite{PGD} &  ResNet-101 & 0.448 & 0.386 & 0.626 & \textbf{0.245} & 0.451 & 1.509 & 0.127 \\
BEVFormer$\dag$~\cite{BEVFormer} & ResNet-101 & 0.535 &0.445 &0.631 &0.257 &0.405 &0.435 &0.143 \\
PolarFormer$\dag$~\cite{jiang2022polarformer} & ResNet-101 & 0.543 &0.457 & 0.612 & 0.257 & 0.392 &0.467 & 0.129\\
PETRv2$\dag$~\cite{petrv2} & ResNet-101 & 0.553 &0.456& 0.601 &0.249& 0.391 &\textbf{0.382} &0.123 \\
 MV2D-T $\ddag$ & ResNet-101 & \textbf{0.573} &\textbf{0.483}&\textbf{0.567}&0.249&\textbf{0.359}&0.395&\textbf{0.116}\\
\midrule

BEVDet~\cite{BEVDet} & Swin-Base & 0.488 & 0.424 & 0.524 & 0.242 & 0.373 & 0.950 & 0.148\\
M2BEV$\ddag$~\cite{M2BEV} & ResNeXt-101 & 0.474 & 0.429 & 0.583 &0.254 &0.376 & 1.053 & 0.190\\
DETR3D$\S$~\cite{DETR3D} & V2-99 & 0.479 & 0.412 & 0.641 & 0.255 & 0.394 & 0.845 & 0.133\\
BEVDet4D\cite{huang2022bevdet4d}  & Swin-Base & 0.569 &0.451& \textbf{0.511}& \textbf{0.241}& 0.386& \textbf{0.301} &0.121 \\
BEVFormer$\S$~\cite{BEVFormer} & V2-99 & 0.569& 0.481 &0.582 &0.256& 0.375& 0.378 &0.126\\
PolarFormer$\S$~\cite{jiang2022polarformer} & V2-99 & 0.572& 0.493 &0.556 &0.256& 0.364& 0.440 &0.127\\
PETRv2$\S$~\cite{PETR} & V2-99 & 0.582 &0.490 &0.561& 0.243& 0.361 &0.343& \textbf{0.120}\\
MV2D-T$\S$ &V2-99& \textbf{0.596} & \textbf{0.511} & 0.525 &0.243 &\textbf{0.357} &0.357 & \textbf{0.120}\\ 
\bottomrule[1.5pt]
\end{tabular}
\caption{
\textbf{3D detection results on nuScenes \texttt{test} set}.
$\dag$: the model is initialized from FCOS3D. 
$\ddag$: the model is pretrained on nuImages. 
$\S$: the model is initialized from DD3D.
}
\label{tab:test_set}
\end{center}
\vspace{-6mm}
\end{table*}

\section{Experiments}
\label{sec:experiments}
\subsection{Datasets and Metrics}
We validate the effectiveness of MV2D on the large-scale nuScenes dataset~\cite{nuscenes2019}.
NuScenes contains 1000 driving sequences, which are divided into 700 samples for training, 150 samples for validation, and 150 samples for testing, respectively.
Each sequence is approximately 20-second long, including annotated 3D bounding boxes from 10 categories of sampled key frames.
Each sample consists of $6$ surround-view images with $1600 \times 900$ resolution, which provide $360^{\circ}$ horizontal FOV in total. We submit \textit{test} set results to the online server for official evaluation. Other experiments are based on the \textit{val} set.

\subsection{Implementation Details}
\label{sec:implementation_details}

\textbf{Training and evaluation.}
All the models are trained using AdamW~\cite{DBLP:conf/iclr/LoshchilovH19} optimizer with a weight decay of $0.01$.
Cosine annealing policy~\cite{loshchilov2016sgdr} is adopted and the initial learning rate is set to $2 \times 10^{-4}$.
Since MV2D generates object queries from 2D detection results, we pretrain the 2D detectors on nuImages~\cite{nuscenes2019} to provide appropriate initialization. Since nuScenes dataset does not provide 2D bounding box annotations, we generate the box labels from 3D bounding boxes following \cite{M2BEV}. We employ image-space augmentation (e.g., gridmask~\cite{chen2020gridmask}, random flip, crop, and resize) and BEV-space augmentation~\cite{BEVDet} during training.
The models are trained for 72 epochs without CBGS~\cite{zhu2019class} in comparison with state-of-the-arts.
No test time augmentation is used during inference.
We denote MV2D-S and MV2D-T as the models without and with multi-frame input respectively. For MV2D-T, we sample two frames from different timestamps with the interval of $15T (T\approx 0.083s)$ in the inference. We use denoise training for MV2D-T following PETRv2~\cite{petrv2} for fair comparison.
Our implementation is based on MMDetection3D~\cite{mmdet3d2020}.

\textbf{Network architecture.}
Faster-RCNN~\cite{FasterRCNN} serves as 2D detector in our experiments.
The 2D score threshold and Non-maximum Suppression (NMS) IoU threshold are set to $0.05$ and $0.6$ respectively.
We use ResNet-50, ResNet-101~\cite{he2016deep} and VoVNetV2~\cite{Vovnet} as backbone networks. 
ResNet-50 and ResNet-101 are equipped with deformable convolution~\cite{dai2017deformable} in the 3rd and 4th stages. 
VoVNetV2 is initialized from a DD3D checkpoint trained with extra data~\cite{DBLP:conf/iccv/ParkAG0G21}. 
The P4 stage in FPN~\cite{lin2017feature} is adopted for dynamic object query generation and relevant feature selection. 
The decoder contains 6 transformer decoder layers, followed by a MLP head for classification and regression.

\begin{figure*}[t!]
	\centering
	\vspace{-3mm}
	\includegraphics[width=0.96\textwidth]{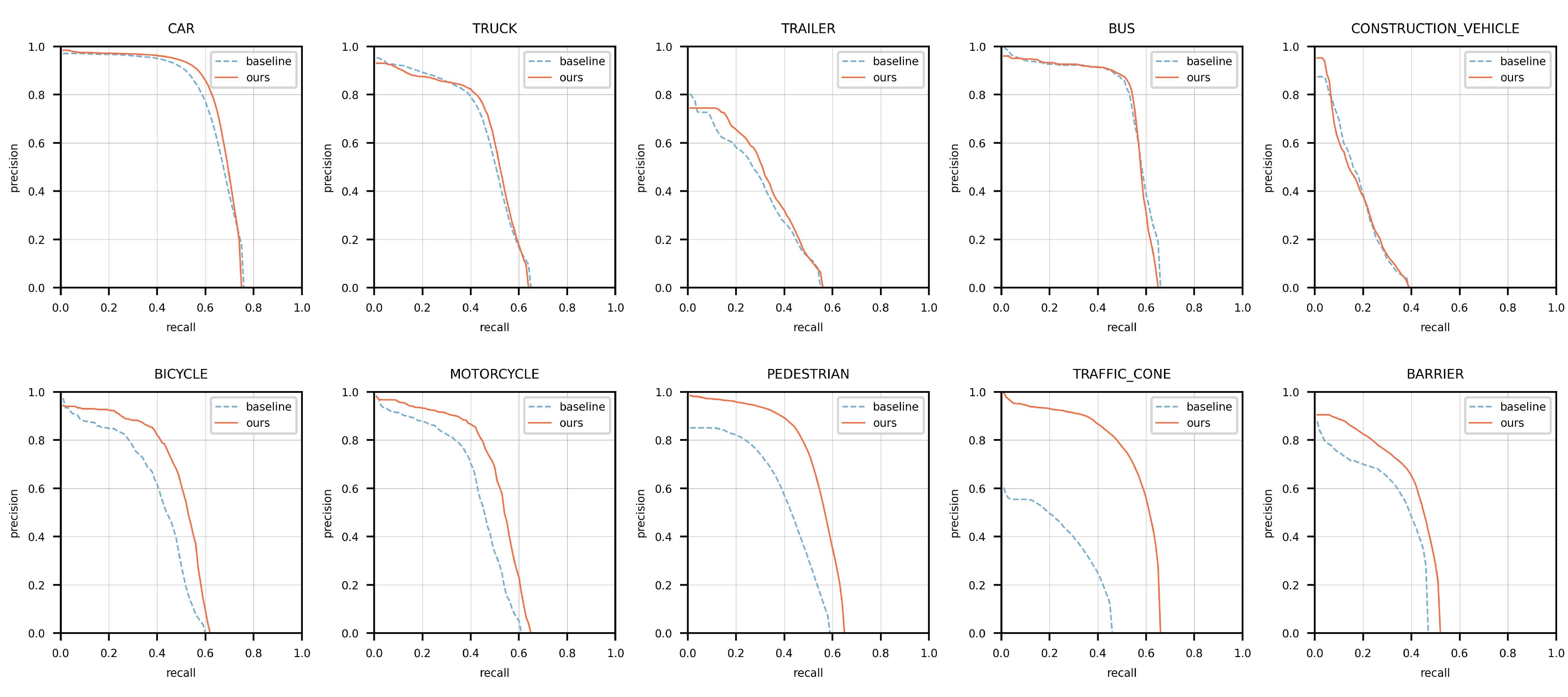}
	\vspace{-2mm}
	\caption{
	Precision-recall curves of different classes under a 2D object detection setting. The models still produce 3D object predictions, while the 2D bounding boxes in each image are generated by projecting 3D bounding boxes using camera parameters. Precision \& recall are calculated with 2D IoU threshold 0.5 in nuScenes \textit{val} set. 
	The blue dash line represents the \textit{baseline} method (fixed object queries). The red solid line represents our method (generated object queries).
	}
	\vspace{-3mm}
	\label{fig:2d_project_comparison}
\end{figure*}

\subsection{Comparison with State-of-the-Arts}
We compare the MV2D performance with state-of-the-art methods on nuScenes \textit{val} set and \textit{test} set. The results are shown in Table~\ref{tab:val_set} and Table~\ref{tab:test_set}.

Table~\ref{tab:val_set} shows the comparison on nuScenes \textit{val} set. 
From the table, MV2D achieves higher performance with both ResNet-50 and ResNet-101. 
In comparison with multi-view 3D object detection methods,
MV2D-T with ResNet-101 achieves $47.1\%$ mAP and $56.1\%$ NDS, which outperforms BEV based methods PolarFormer~\cite{jiang2022polarformer} by $3.9\%$ mAP and $3.3\%$ NDS.
In comparison with query based methods, MV2D-T improves PETRv2 \cite{PETR} by $5.0\%$ and $3.7\%$ on mAP and NDS. Compared to other methods that also use single frame input, MV2D-S shows a consistent advantage on mAP.
The apparent improvement on mAP suggests the 3D object localization ability can be enhanced by introducing 2D detection.  

Table~\ref{tab:test_set} shows the comparison on nuScenes \textit{test} set.
As in the table, MV2D with ResNet-101 achieves $48.3\%$ mAP and $57.3\%$ NDS, which outperforms other methods by $2.6\%$ mAP and $2.0\%$ NDS.
MV2D with VoVNetV2 achieves $51.1\%$ mAP and $59.6\%$ NDS, delivering better performance compared to existing methods.

\begin{table}[t!]
\begin{center}
\resizebox{0.47\textwidth}{!}{

\begin{tabular}{cccc|c|c c}
\toprule[1.pt]
 \# & query & curr. K\&V & hist. K\&V & mAP$\uparrow$  & NDS$\uparrow$ \\
 \midrule
 1& fixed (900)  & all & - & 36.2& 39.9 \\
 2& fixed (300)& all & - & 36.1& 39.4 \\
 3& fixed (1500)& all & - & 36.6& 40.6 \\
 4& dynamic  & all & - & 38.0& 41.2 \\
 5& dynamic & Top1 & - & 38.2 & 42.4 \\
 6& dynamic & AO & -& 38.4 & 42.6 \\
 \hline
 7 & fixed (900) & all & all& 38.6& 47.4\\
 8& dynamic  & AO & AO &41.4 & 51.1\\

\bottomrule[1.pt]
\end{tabular}
}
\caption{
Ablation study on the object query and key\&value in the cross attention layers.
The number of generated queries is calculated on average. 
\textit{curr.} means current frame, \textit{hist.} means history frame, \textit{Top1} means top1 IoU, \textit{AO} means all overlapped.
}
\label{tab:abl_1}
\end{center}
\vspace{-6mm}
\end{table}

\subsection{Ablation Study}
In this section, we provide a detailed analysis of the core design choices of MV2D. All the experiments are based on ResNet-50 backbone, 1408$\times$512 input resolution. The models are trained for 24 epochs.
For fair comparison, we design a baseline method also trained with 2D detection loss and initialized from a nuImages pretrained checkpoint. This method uses a fixed number of learnable object queries to aggregate information from the whole multi-view feature maps as in PETR~\cite{PETR}.  
As in Table~\ref{tab:abl_1}, we denote the design in $\#1$ which follows PETR to use 900 object queries as \textit{baseline} in our experiments.

\begin{table}[t!]
\begin{center}
\resizebox{0.37\textwidth}{!}{
\setlength{\tabcolsep}{1.9mm}
\begin{tabular}{cc|c c}
\toprule[1.pt]
\#& object query generation & mAP$\uparrow$  & NDS$\uparrow$ \\
 \midrule
1&uniform sampling & 37.1& 40.1\\
2&scale based depth  & 37.3& 40.8\\
3&query generator & 38.4& 42.6\\

\bottomrule[1.pt]
\end{tabular}
}
\caption{
Ablation study on the choices of object query generation. We compare the query generator with two alternatives, ``uniform sampling'' and ``scale based depth''.
}
\label{tab:query_generator}
\end{center}
\vspace{-6mm}
\end{table}

\textbf{Generated object queries vs. fixed object queries.}
We first compare the generated object queries with fixed object queries.  
From $\#1$ to $\#3$ in Table~\ref{tab:abl_1}, it can be observed that in the case of fixed queries, a larger query amount can improve performance since the fixed query based methods rely on densely placed object queries to localize objects. 
According to $\#4$, when replacing the fixed queries by dynamically generated queries, mAP and NDS are improved by $1.8\%$ and $1.3\%$ respectively. 
This demonstrates the object queries generated by 2D detector produce higher-quality object location hypotheses benefiting 3D detection.

\textbf{Effectiveness of relevant features.}
We also validate the effectiveness of using only relevant features for object query updating.
In Table~\ref{tab:abl_1}, $\#4$ represents using the whole multi-view feature maps for each object query, $\#5$ and  $\#6$  represent using only the relevant object features for each object query, as described in Section~\ref{sec:correlated_feature_selection}. 
As shown by the results, using only the relevant object features selected by the ``top1 IoU`` rule for object query updating can improve mAP and NDS by $0.2\%$ and $1.2\%$ respectively. The ``all overlapped`` rule can bring an additional $0.2\%$ gain for both mAP and NDS.
In the experiments, we implement the relevant object feature selection based on ``all overlapped'' rule due to its better performance.
This result verifies the effectiveness of object query to aggregate information in a specified foreground region.

\textbf{History information.}
We  sample one additional  camera sweep to provide history information for MV2D. As shown in Table~\ref{tab:abl_1}, when using multi-frame input, $\#8$ improves over the single-frame version $\#6$ by $3.0\%$ mAP and $8.5\%$ NDS. For fair comparison, $\#8$ also improves the \textit{baseline} with history information $\#7$ by $2.8\%$ mAP and $3.7\%$ NDS.

\begin{table}[t!]
\begin{center}
\resizebox{0.45\textwidth}{!}{

\begin{tabular}{ccc|c c}
\toprule[1.pt]
resolution&  backbone &2D detector  & mAP$\uparrow$  & NDS$\uparrow$ \\
 \midrule
 1408$\times$512&R50 &YOLOX  & 41.3& 50.0\\
 1408$\times$512&R50 &RetinaNet & 41.1& 50.9\\
 1408$\times$512&R50 &Faster R-CNN  & 41.4& 51.1\\

\bottomrule[1.pt]
\end{tabular}
}
\caption{
Experiments with different 2D detectors. 
}
\label{tab:abl_detectors}
\end{center}
\vspace{-5mm}
\end{table}
  
\begin{table}[t!]
\begin{center}
\resizebox{0.38\textwidth}{!}{

\begin{tabular}{c|c|cc}
\toprule[1.0pt]
 method & cross-attention mode & mAP  & NDS \\
 \midrule
 MV2D& deformable attention  & 39.7 & 48.4 \\
 MV2D& relevant object region & 41.4 & 51.1\\
\bottomrule[1.0pt]
\end{tabular}
}
\caption{Comparison of cross-attention mode.}
\label{tab:comparison_cross_attn}
\end{center}
\vspace{-5mm}
\end{table}

\begin{table}[t!]
\begin{center}
\resizebox{0.48\textwidth}{!}{

\begin{tabular}{c|ccc|ccc}
\toprule[1pt]
 method & pre-trained & mAP  & NDS & pre-trained & mAP  & NDS\\
 \midrule
 baseline& COCO&37.0 & 45.6& nuImages  & 38.6 & 47.4 \\
 MV2D& COCO& 39.3& 49.8& nuImages& 41.4 & 51.1\\
\bottomrule[1pt]
\end{tabular}
}
\caption{Comparison of pre-trained weights.}
\label{tab:comparison_pretrained}
\end{center}
\vspace{-5mm}
\end{table}

\textbf{Design of object query generator.}
Except for the design described in Section~\ref{sec:query_generator}, we also experiment with other alternatives for object query generator. One alternative is to sample 10 depth values uniformly from [0.5m, 65m] for each RoI, named ``uniform sampling''. Another is to derive depth value based on apparent object scale in image~\cite{depth_in_single_images}, named ``scale based depth''.  The experiments are based on single frame setting and the results are listed in Table~\ref{tab:query_generator}. The two alternatives ``uniform sampling'' and ``scale based depth'' assume simpler distributions of the depth value of objects, hindering the 3D object localization ability.

\textbf{Comparison with deformable attention.}
  Deformable Attention~\cite{Deformable-DETR} also limits the attention computation within the local scope of relevant object features. However, Deformable attention cannot guarantee that its generated sampling points will cover the corresponding object area. Our method explicitly imposes this constraint, thereby yielding superior results as shown in Table~\ref{tab:comparison_cross_attn}.

\textbf{Ablation of pre-trained weights:}
To evaluate the impact of pre-training, we provide the comparison in Table~\ref{tab:comparison_pretrained}. The baseline corresponds to the setting of \#7 in Table~\ref{tab:abl_1}. Our method has consistent performance improvements with different pretrained weights.

\subsection{Generalizability on Different 2D Detectors}
We provide experiments with different 2D detectors to validate the generalizability of MV2D.
We choose 3 kinds of 2D detectors, including a two-stage detector Faster R-CNN, a single-stage anchor-based  detector RetinaNet~\cite{RetinaNet} and a single-stage anchor-free detector YOLOX~\cite{yolox2021}. All the models are trained for 24 epochs. The detailed experiment setting is in supplementary materials.

As shown in Table~\ref{tab:abl_detectors}, with a stronger object detector Faster R-CNN, MV2D can achieve higher performance due to better object proposal quality. If equipped with a lightweight object detector such as RetinaNet and YOLOX, MV2D also exhibits decent performance, demonstrating the generalizability of MV2D.

\begin{figure}[t!]
	\centering
	\vspace{-3mm}
	\includegraphics[width=0.4\textwidth]{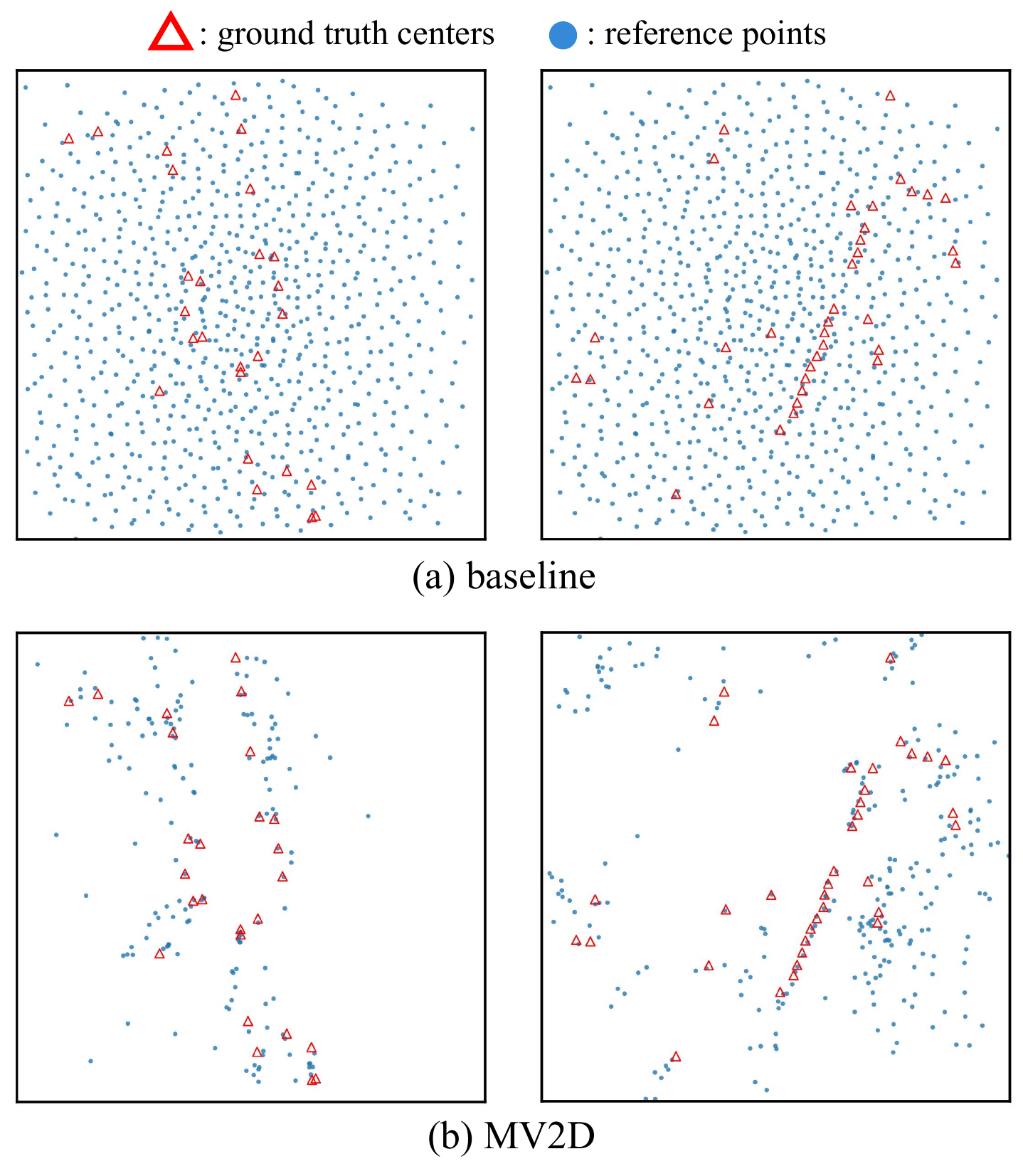}
	\vspace{-1mm}
	\caption{Illustration of different kinds of object queries. The top row (a) represents the \textit{baseline} method (fixed object queries) and the bottom row (b) represents our method (generated object queries). We visualize ground truth center locations and the reference points in BEV space. The samples are taken from nuScenes \textit{val} set. We suggest readers view this figure by zooming in.
	}
	\vspace{-4mm}
	\label{fig:ref_points}
\end{figure}

\subsection{Qualitative Analysis}

In this section, we analyze how might a 2D detector promote the performance of multi-view 3D object detection.
To verify if the object queries generated from 2D detector can better localize objects with the help of image semantics, we compare the 2D object detection performance of the \textit{baseline} method (fixed object queries) and our method (generated object queries) by perspective projection. 
Specifically, we project the 3D bounding boxes into each image, then calculate the minimum 2D bounding boxes as 2D results. The performance is evaluated between the projected 2D bounding boxes of predictions and ground truths. We set the IoU threshold to 0.5 for true positive assignment. 

In Figure~\ref{fig:2d_project_comparison}, we draw the precision-recall curves of 10 classes in nuScenes. As seen from the figure, our method achieves higher precision and recall compared to the \textit{baseline} method, especially for the classes with relatively smaller sizes.
This demonstrates that 2D object detectors can make the most of image semantics to discover objects from image space, which can serve as solid evidence of object existence to be exploited by 3D detectors.
For complementary, we provide the visualization of different kinds of queries in Figure~\ref{fig:ref_points}. Though the generated queries are much sparser compared to the fixed queries, they are mostly distributed around the objects.
More visualizations and case studies are in supplementary materials.

\section{Conclusion}
In this paper, we propose Multi-View 2D Objects guided 3D Object Detector (MV2D) for multi-view 3D object detection. In our framework, we utilize 2D objects as sparse queries and adopt a sparse cross attention module to constrain the information aggregation. In our experiments, we demonstrate promising results on nuScenes dataset with our proposed MV2D framework. MV2D can lift any 2D detector to 3D detection, and we believe the insight of utilizing the 2D objects as guidance can further inspire the design of multi-view 3D object detection methods.

\textbf{Limitations.}
Since MV2D generates object queries from 2D detections, an object might be missed if the 2D detector fails to detect it in all the camera views. 

\section*{Acknowledgement}
This work was supported in part by the National Key R\&D Program of China under Grant 2022ZD0115502, in part by the National Natural Science Foundation of China under Grant 62122010, and in part by the Fundamental Research Funds for the Central Universities. 

{\small
\bibliographystyle{ieee_fullname}
\bibliography{egbib}

\begin{thebibliography}{10}\itemsep=-1pt

\bibitem{DBLP:conf/iccv/Brazil019}
Garrick Brazil and Xiaoming Liu.
\newblock {M3D-RPN:} monocular 3d region proposal network for object detection.
\newblock In {\em {ICCV}}, 2019.

\bibitem{nuscenes2019}
Holger Caesar, Varun Bankiti, Alex~H. Lang, Sourabh Vora, Venice~Erin Liong, Qiang Xu, Anush Krishnan, Yu Pan, Giancarlo Baldan, and Oscar Beijbom.
\newblock nuscenes: {A} multimodal dataset for autonomous driving.
\newblock In {\em {CVPR}}, 2020.

\bibitem{DETR}
Nicolas Carion, Francisco Massa, Gabriel Synnaeve, Nicolas Usunier, Alexander Kirillov, and Sergey Zagoruyko.
\newblock End-to-end object detection with transformers.
\newblock In {\em {ECCV}}, 2020.

\bibitem{chen2020gridmask}
Pengguang Chen, Shu Liu, Hengshuang Zhao, and Jiaya Jia.
\newblock Gridmask data augmentation.
\newblock {\em arXiv preprint arXiv:2001.04086}, 2020.

\bibitem{PolarDETR}
Shaoyu Chen, Xinggang Wang, Tianheng Cheng, Qian Zhang, Chang Huang, and Wenyu Liu.
\newblock Polar parametrization for vision-based surround-view 3d detection.
\newblock {\em arXiv preprint arXiv:2206.10965}, 2022.

\bibitem{mmdet3d2020}
MMDetection3D Contributors.
\newblock {MMDetection3D: OpenMMLab} next-generation platform for general {3D} object detection.
\newblock \url{https://github.com/open-mmlab/mmdetection3d}, 2020.

\bibitem{dai2017deformable}
Jifeng Dai, Haozhi Qi, Yuwen Xiong, Yi Li, Guodong Zhang, Han Hu, and Yichen Wei.
\newblock Deformable convolutional networks.
\newblock In {\em ICCV}, 2017.

\bibitem{depth_in_single_images}
Tom~van Dijk and Guido~de Croon.
\newblock How do neural networks see depth in single images?
\newblock In {\em Proceedings of the IEEE/CVF International Conference on Computer Vision}, pages 2183--2191, 2019.

\bibitem{FSD}
Lue Fan, Feng Wang, Naiyan Wang, and Zhaoxiang Zhang.
\newblock Fully sparse 3d object detection.
\newblock In {\em NeurIPS}, 2022.

\bibitem{yolox2021}
Zheng Ge, Songtao Liu, Feng Wang, Zeming Li, and Jian Sun.
\newblock Yolox: Exceeding yolo series in 2021.
\newblock {\em arXiv preprint arXiv:2107.08430}, 2021.

\bibitem{FastRCNN}
Ross Girshick.
\newblock {Fast R-CNN}.
\newblock In {\em ICCV}, 2015.

\bibitem{RCNN}
Ross Girshick, Jeff Donahue, Trevor Darrell, and Jitendra Malik.
\newblock Rich feature hierarchies for accurate object detection and semantic segmentation.
\newblock In {\em CVPR}, 2014.

\bibitem{MaskRCNN}
Kaiming He, Georgia Gkioxari, Piotr Doll{\'a}r, and Ross Girshick.
\newblock {Mask R-CNN}.
\newblock In {\em ICCV}, 2017.

\bibitem{he2016deep}
Kaiming He, Xiangyu Zhang, Shaoqing Ren, and Jian Sun.
\newblock Deep residual learning for image recognition.
\newblock In {\em CVPR}, 2016.

\bibitem{huang2022bevdet4d}
Junjie Huang and Guan Huang.
\newblock Bevdet4d: Exploit temporal cues in multi-camera 3d object detection.
\newblock {\em arXiv preprint arXiv:2203.17054}, 2022.

\bibitem{BEVDet}
Junjie Huang, Guan Huang, Zheng Zhu, and Dalong Du.
\newblock {BEVD}et: High-performance multi-camera 3d object detection in bird-eye-view.
\newblock {\em arXiv preprint arXiv:2112.11790}, 2021.

\bibitem{jiang2022polarformer}
Yanqin Jiang, Li Zhang, Zhenwei Miao, Xiatian Zhu, Jin Gao, Weiming Hu, and Yu-Gang Jiang.
\newblock Polarformer: Multi-camera 3d object detection with polar transformers.
\newblock {\em arXiv preprint arXiv:2206.15398}, 2022.

\bibitem{DBLP:conf/cvpr/KuPW19}
Jason Ku, Alex~D. Pon, and Steven~L. Waslander.
\newblock Monocular 3d object detection leveraging accurate proposals and shape reconstruction.
\newblock In {\em {CVPR}}, 2019.

\bibitem{kuhn1955hungarian}
Harold~W Kuhn.
\newblock The hungarian method for the assignment problem.
\newblock {\em Naval research logistics quarterly}, 1955.

\bibitem{Vovnet}
Youngwan Lee, Joong{-}Won Hwang, Sangrok Lee, Yuseok Bae, and Jongyoul Park.
\newblock An energy and gpu-computation efficient backbone network for real-time object detection.
\newblock In {\em {CVPRW}}, 2019.

\bibitem{UVTR}
Yanwei Li, Yilun Chen, Xiaojuan Qi, Zeming Li, Jian Sun, and Jiaya Jia.
\newblock Unifying voxel-based representation with transformer for 3d object detection.
\newblock In {\em {NeurIPS}}, 2022.

\bibitem{BEVDepth}
Yinhao Li, Zheng Ge, Guanyi Yu, Jinrong Yang, Zengran Wang, Yukang Shi, Jianjian Sun, and Zeming Li.
\newblock {BEVD}epth: Acquisition of reliable depth for multi-view 3d object detection.
\newblock {\em arXiv preprint arXiv:2206.10092}, 2022.

\bibitem{BEVFormer}
Zhiqi Li, Wenhai Wang, Hongyang Li, Enze Xie, Chonghao Sima, Tong Lu, Qiao Yu, and Jifeng Dai.
\newblock {BEVF}ormer: Learning bird's-eye-view representation from multi-camera images via spatiotemporal transformers.
\newblock In {\em ECCV}, 2022.

\bibitem{lin2017feature}
Tsung-Yi Lin, Piotr Doll{\'a}r, Ross Girshick, Kaiming He, Bharath Hariharan, and Serge Belongie.
\newblock Feature pyramid networks for object detection.
\newblock In {\em CVPR}, 2017.

\bibitem{RetinaNet}
Tsung-Yi Lin, Priya Goyal, Ross Girshick, Kaiming He, and Piotr Doll{\'a}r.
\newblock Focal loss for dense object detection.
\newblock In {\em ICCV}, 2017.

\bibitem{PETR}
Yingfei Liu, Tiancai Wang, Xiangyu Zhang, and Jian Sun.
\newblock {PETR}: Position embedding transformation for multi-view 3d object detection.
\newblock In {\em ECCV}, 2022.

\bibitem{petrv2}
Yingfei Liu, Junjie Yan, Fan Jia, Shuailin Li, Qi Gao, Tiancai Wang, Xiangyu Zhang, and Jian Sun.
\newblock Petrv2: A unified framework for 3d perception from multi-camera images.
\newblock {\em arXiv preprint arXiv:2206.01256}, 2022.

\bibitem{loshchilov2016sgdr}
Ilya Loshchilov and Frank Hutter.
\newblock {SGDR:} stochastic gradient descent with warm restarts.
\newblock In {\em {ICLR}}, 2017.

\bibitem{DBLP:conf/iclr/LoshchilovH19}
Ilya Loshchilov and Frank Hutter.
\newblock Decoupled weight decay regularization.
\newblock In {\em ICLR}, 2019.

\bibitem{mousavian20173d}
Arsalan Mousavian, Dragomir Anguelov, John Flynn, and Jana Kosecka.
\newblock 3d bounding box estimation using deep learning and geometry.
\newblock In {\em CVPR}, 2017.

\bibitem{DBLP:conf/iccv/ParkAG0G21}
Dennis Park, Rares Ambrus, Vitor Guizilini, Jie Li, and Adrien Gaidon.
\newblock Is pseudo-lidar needed for monocular 3d object detection?
\newblock In {\em ICCV}, 2021.

\bibitem{LSS}
Jonah Philion and Sanja Fidler.
\newblock Lift, splat, shoot: Encoding images from arbitrary camera rigs by implicitly unprojecting to 3d.
\newblock In {\em {ECCV}}, 2020.

\bibitem{YOLO}
Joseph Redmon, Santosh Divvala, Ross Girshick, and Ali Farhadi.
\newblock You only look once: Unified, real-time object detection.
\newblock In {\em CVPR}, 2016.

\bibitem{FasterRCNN}
Shaoqing Ren, Kaiming He, Ross Girshick, and Jian Sun.
\newblock {Faster R-CNN}: Towards real-time object detection with region proposal networks.
\newblock In {\em {NeurIPS}}, 2015.

\bibitem{ImVoxelNet}
Danila Rukhovich, Anna Vorontsova, and Anton Konushin.
\newblock Imvoxelnet: Image to voxels projection for monocular and multi-view general-purpose 3d object detection.
\newblock In {\em {WACV}}, 2022.

\bibitem{DBLP:conf/iccv/SimonelliBPLK19}
Andrea Simonelli, Samuel~Rota Bul{\`{o}}, Lorenzo Porzi, Manuel Lopez{-}Antequera, and Peter Kontschieder.
\newblock Disentangling monocular 3d object detection.
\newblock In {\em {ICCV}}, 2019.

\bibitem{FCOS}
Zhi Tian, Chunhua Shen, Hao Chen, and Tong He.
\newblock {FCOS}: Fully convolutional one-stage object detection.
\newblock In {\em ICCV}, 2019.

\bibitem{vaswani2017attention}
Ashish Vaswani, Noam Shazeer, Niki Parmar, Jakob Uszkoreit, Llion Jones, Aidan~N Gomez, {\L}ukasz Kaiser, and Illia Polosukhin.
\newblock Attention is all you need.
\newblock In {\em NeurIPS}, 2017.

\bibitem{FCOS3D}
Tai Wang, Xinge Zhu, Jiangmiao Pang, and Dahua Lin.
\newblock {FCOS3D:} fully convolutional one-stage monocular 3d object detection.
\newblock In {\em {ICCVW}}, 2021.

\bibitem{PGD}
Tai Wang, Xinge Zhu, Jiangmiao Pang, and Dahua Lin.
\newblock Probabilistic and geometric depth: Detecting objects in perspective.
\newblock In {\em {CoRL}}, 2021.

\bibitem{DETR3D}
Yue Wang, Vitor Guizilini, Tianyuan Zhang, Yilun Wang, Hang Zhao, and Justin Solomon.
\newblock {DETR3D:} 3d object detection from multi-view images via 3d-to-2d queries.
\newblock In {\em CoRL}, 2021.

\bibitem{M2BEV}
Enze Xie, Zhiding Yu, Daquan Zhou, Jonah Philion, Anima Anandkumar, Sanja Fidler, Ping Luo, and Jose~M Alvarez.
\newblock {M\textsuperscript{2}BEV}: Multi-camera joint 3d detection and segmentation with unified birds-eye view representation.
\newblock {\em arXiv preprint arXiv:2204.05088}, 2022.

\bibitem{efficientdetr}
Zhuyu Yao, Jiangbo Ai, Boxun Li, and Chi Zhang.
\newblock Efficient detr: improving end-to-end object detector with dense prior.
\newblock {\em arXiv preprint arXiv:2104.01318}, 2021.

\bibitem{CVT}
Brady Zhou and Philipp Kr{\"a}henb{\"u}hl.
\newblock Cross-view transformers for real-time map-view semantic segmentation.
\newblock In {\em CVPR}, 2022.

\bibitem{DBLP:journals/corr/abs-1904-07850}
Xingyi Zhou, Dequan Wang, and Philipp Kr{\"{a}}henb{\"{u}}hl.
\newblock Objects as points.
\newblock {\em CoRR}, abs/1904.07850, 2019.

\bibitem{zhu2019class}
Benjin Zhu, Zhengkai Jiang, Xiangxin Zhou, Zeming Li, and Gang Yu.
\newblock Class-balanced grouping and sampling for point cloud 3d object detection.
\newblock {\em arXiv preprint arXiv:1908.09492}, 2019.

\bibitem{Deformable-DETR}
Xizhou Zhu, Weijie Su, Lewei Lu, Bin Li, Xiaogang Wang, and Jifeng Dai.
\newblock Deformable {DETR:} deformable transformers for end-to-end object detection.
\newblock In {\em {ICLR}}, 2021.

\end{thebibliography}
}

\appendix
\clearpage

\section{Experiments with More Architectures}
We provide more experiments with different 2D detectors and feature extractors in this section.
The 2D detector part of MV2D is pretrained on nuImages~\cite{nuscenes2019}, then the 2D detector part and 3D detector part is jointly trained on nuScenes \textit{train} set. 3D object detection performance and model latency are evaluated on nuScenes \textit{val} set~\cite{nuscenes2019}.
All the models are trained for 24 epochs without CBGS.
For model latency, we only consider the latency of network forward pass and ignore the pre-processing and post-processing time (e.g., image loading and format converting).
The latency is evaluated on a single NVIDIA RTX 3090 GPU with batch size 1.
We provide the detailed model architectures below.

\subsection{Model Architecture}
\paragraph{2D detector}
Without loss of generality, we choose 3 kinds of 2D detectors, including Faster R-CNN~\cite{FasterRCNN}, a single-stage anchor-based 2D detector RetinaNet~\cite{RetinaNet} and a single-stage anchor-free 2D detector YOLOX~\cite{yolox2021}.

\paragraph{Feature pyramid}
For models with ResNet-50 backbone and Faster R-CNN detector, the feature pyramid is built to produce feature maps with downsample stride $\{4, 8, 16, 32, 64\}$.
For models with ResNet-50 backbone and RetinaNet/YOLOX detector, the feature pyramid is built to produce feature maps with downsample stride $\{8, 16, 32, 64,128\}$.

\paragraph{Decoder layers} The decoder in MV2D contains 6 decoder layers by default. We also experiment with different numbers of decoder layers.

\subsection{Performance Comparison}
We equip MV2D with different 2D detectors and feature extractors, then evaluate their latency and performance on nuScenes \textit{val} set. The results are listed in Table~\ref{tab:comprehensive}. 
As demonstrated by the results, under 1408$\times$512 input resolution, MV2D with Faster R-CNN as 2D detector achieves the highest performance of $41.4\%$ mAP and $51.1\%$ NDS with the inference latency of 380ms. When using RetinaNet as 2D detector, the performance drops slightly, obtaining $41.1\%$ mAP and $50.9\%$ NDS. With a faster single-stage detector YOLOX, the inference latency is reduced to 246ms with a decent performance of $41.3\%$ mAP and $50.0\%$ NDS.
These results suggest that MV2D can adapt to different 2D detectors and a very lightweight 2D detector can still work.
Under a smaller input resolution of 800$\times$320, MV2D with YOLOX achieves $37.7\%$ mAP and $47.0\%$ NDS and reduces the latency to 115ms.
These experiments show that MV2D can generalize well to other architectures.

\begin{figure}[t!]
	\centering
	\includegraphics[width=0.47\textwidth]{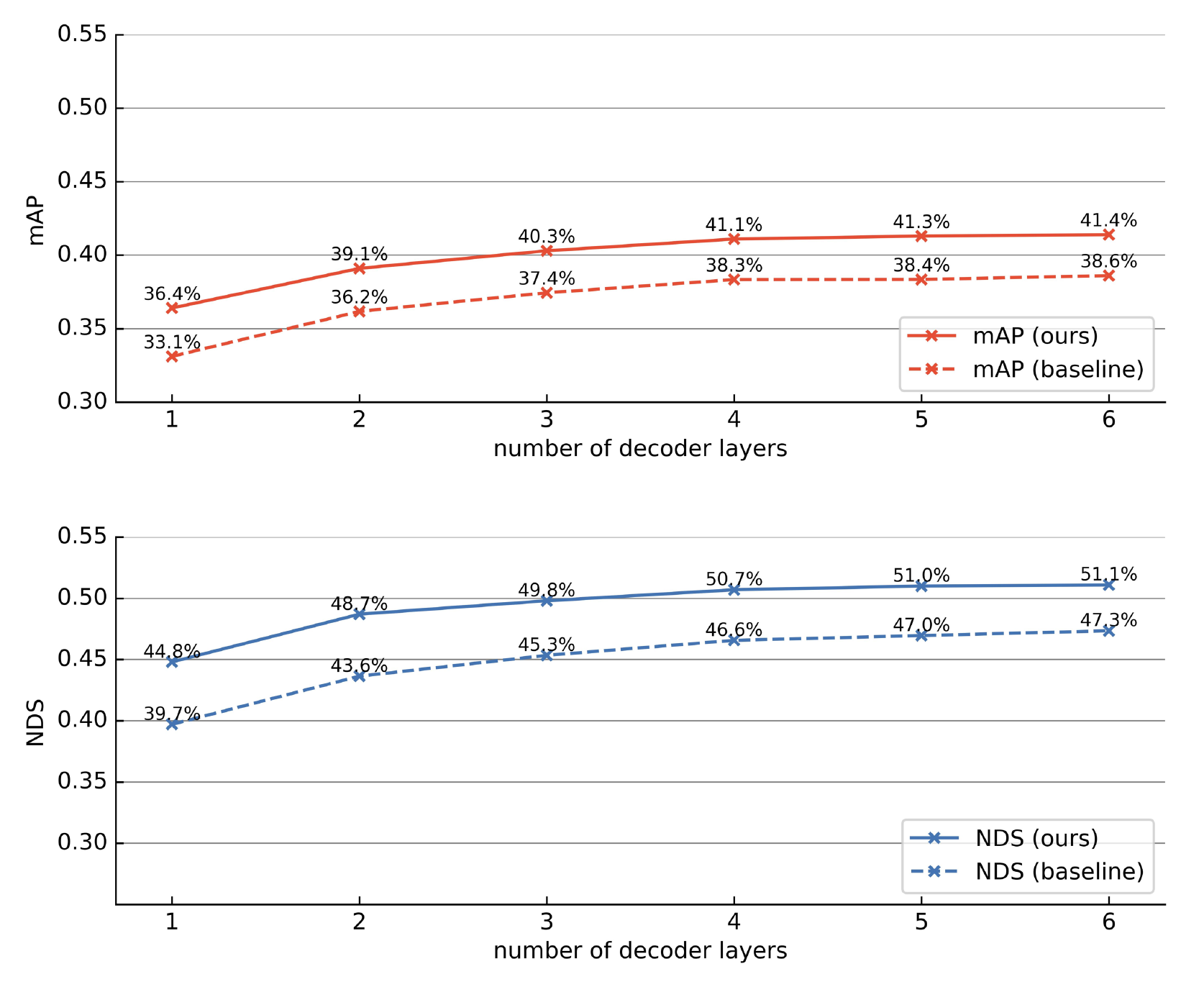}
	\vspace{-2mm}
	\caption{Comparison on different decoder layers.}
	\vspace{-6mm}
	\label{fig:stats_decoder_layers}
\end{figure}

    In Figure~\ref{fig:stats_decoder_layers}, we evaluate the performance of different numbers of decoder layers. MV2D is based on Faster R-CNN as 2D detector with 1408$\times$512 input resolution, and the baseline method is based on fixed object queries with the same input resolution.
 With 1 decoder layer, MV2D achieves $36.4\%$ mAP and $44.8\%$ NDS. With 2 decoder layers, mAP and NDS improve by $2.7\%$ and $3.9\%$ respectively. As the number of decoder layers increases, the mAP and NDS also increase.
 It can be seen that MV2D with 2 decoder layers outperforms the baseline method with 6 decoder layers on both mAP and NDS.


\begin{table*}[tb]
\begin{center}
\begin{tabular}{c c c |c| c c | c c c c c }
\toprule[1.5pt]
Resolution & Backbone & 2D Detector & Latency & mAP  & NDS & mATE     & mASE     & mAOE & mAVE & mAAE\\

\midrule
800$\times$320 & ResNet-50 & YOLOX & 115ms & 0.377 & 0.470 &0.737 & 0.280& 0.532& 0.427 &0.213 \\
1408$\times$512 & ResNet-50 &  YOLOX& 246ms & 0.413& 0.500 &0.697 & 0.271 & 0.496 & 0.402 &  0.203 \\
1408$\times$512 & ResNet-50 &  RetinaNet  & 368ms&  0.411& 0.509 & 0.696 & 0.272 & 0.418 & 0.387 &0.185\\
1408$\times$512 & ResNet-50 &  Faster R-CNN & 380ms &0.414& 0.511 & 0.694 & 0.272 & 0.427 & 0.396 & 0.172\\

\bottomrule[1.5pt]
\end{tabular}
\caption{
Performance comparison on nuScenes \textit{val} set. 
}
\label{tab:comprehensive}
\end{center}
\end{table*}

\begin{figure*}[t!]
	\centering
	\includegraphics[width=0.92\textwidth]{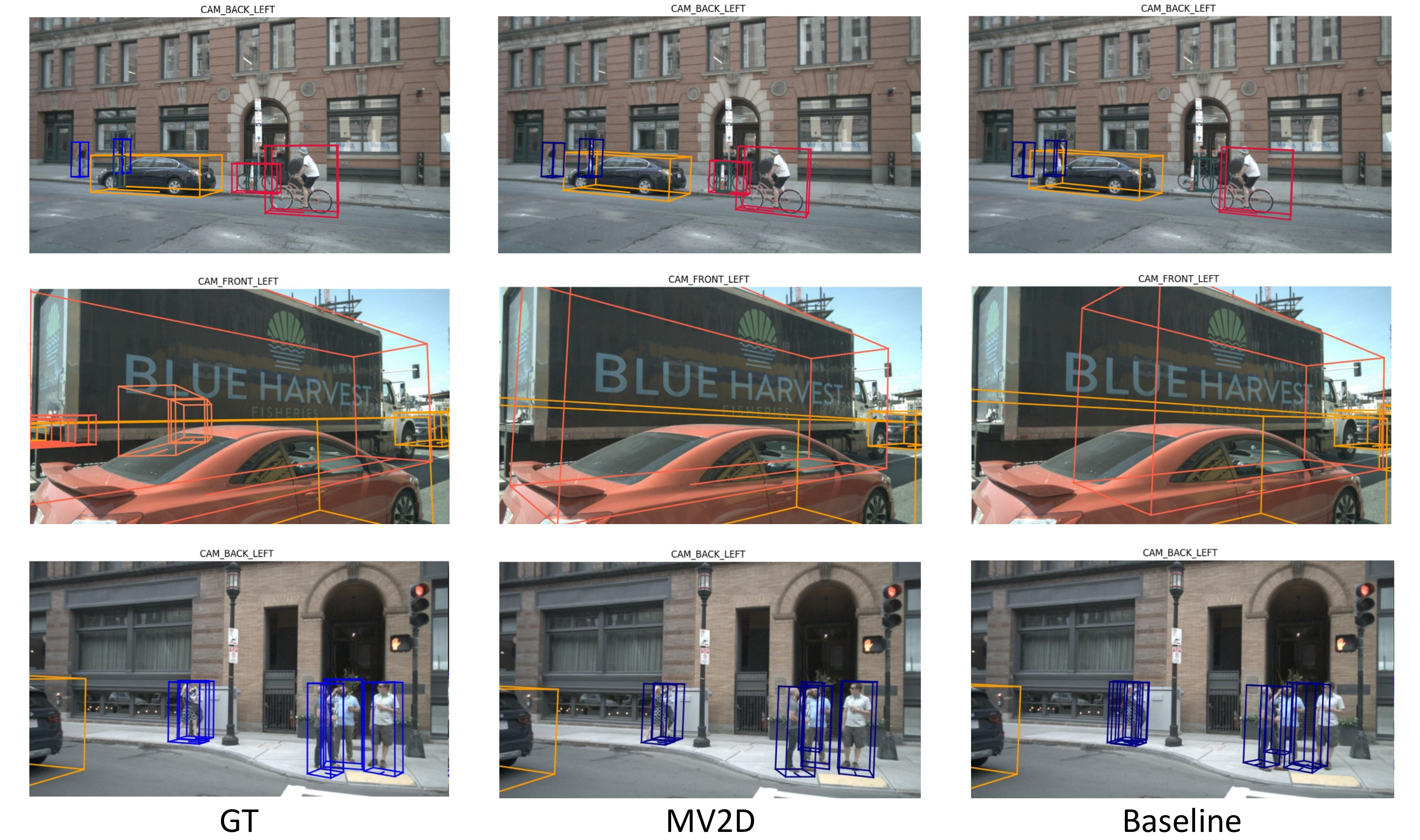}
	\caption{Some qualitative results.}
	\label{fig:qualitative_compare}
\end{figure*}

\begin{figure*}[t!]
	\centering
	\includegraphics[width=0.72\textwidth]{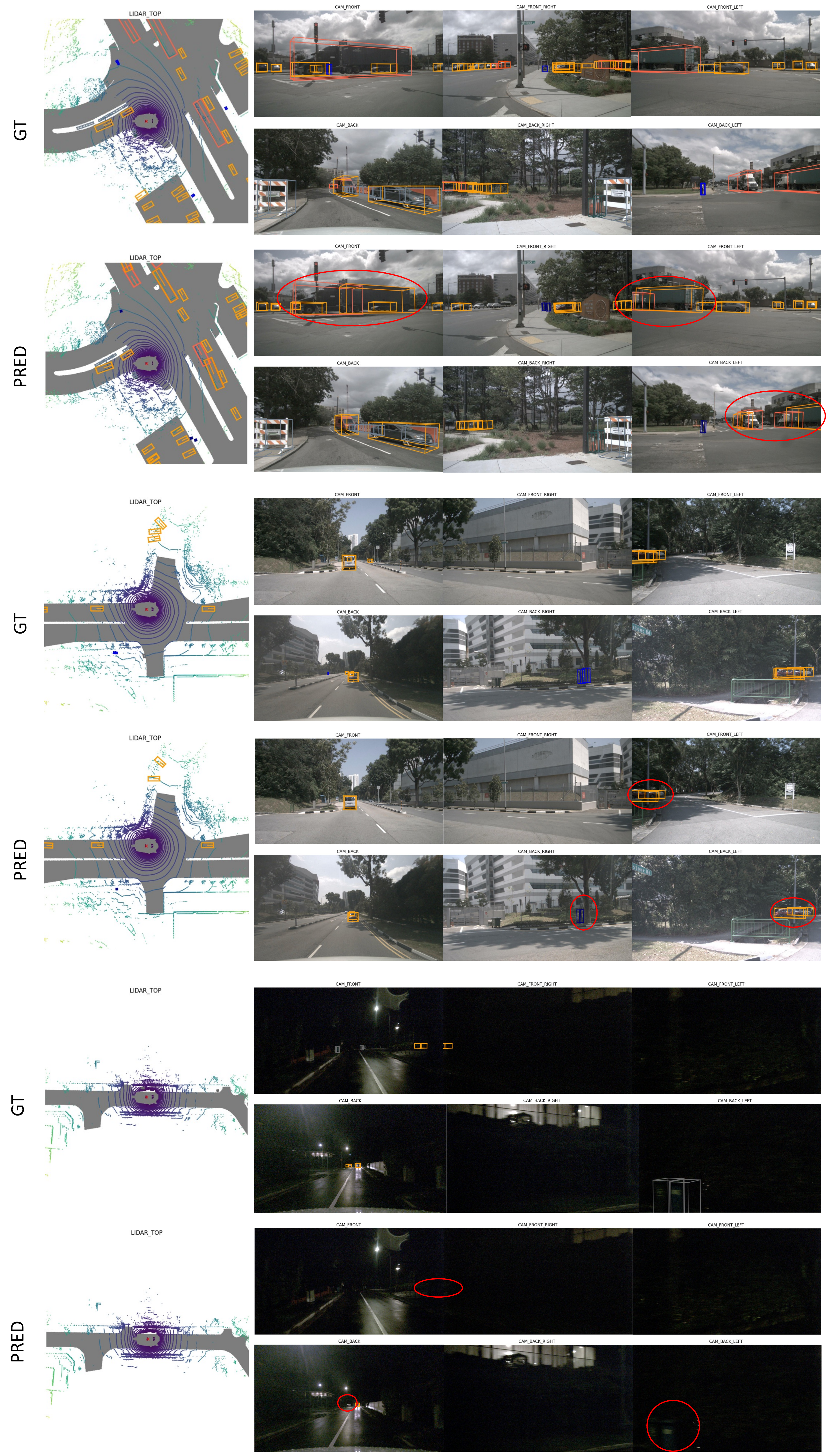}
	\caption{Illustration of some failure cases.}
	\label{fig:failure_case}
\end{figure*}

\section{More Visualizations}

\subsection{Qualitative Results}
We compare MV2D with the baseline method (using fixed object queries and gathering information from all image regions) and show the qualitative results.
The 3D object detection results are illustrated in Figure~\ref{fig:qualitative_compare}.
Row 1 to row 3 is drawn from different data samples. Line 1 to line 3 represent ground truth, the MV2D detection results, and the baseline detection results, respectively. 
The baseline method might fail to detect (the bicycle in row 1) or mislocate (the truck in row 2 and the persons in row 3) some objects in 3D space. However,  most of these objects can be detected by a 2D detector in the image space.
Thus 2D detection can provide rich evidence about object existence and location. By exploiting this information, MV2D can generate more accurate 3D detection results.

\subsection{Failure Case Analysis}
We also analyze the failure case of MV2D. Some examples are shown in Figure~\ref{fig:failure_case}.
From row 1 to row 2, MV2D sometimes splits a  ``truck'' object into a ``truck'' object and a ``trailer'' object.
From row 3 to row 4, if objects are heavily occluded, the 2D detector might fail to detect them successfully, causing false negatives in MV2D.  
From row 5 to row 6, if there are extreme lighting conditions or large motion blurs, the 2D detector can also fail to detect some objects and impair the performance of MV2D.

\end{document}